\documentclass{article}

\usepackage{PRIMEarxiv}

\usepackage[utf8]{inputenc} 
\usepackage[T1]{fontenc}    
\usepackage{hyperref}       
\usepackage{url}            
\usepackage{booktabs}       
\usepackage{amsfonts}       
\usepackage{nicefrac}       
\usepackage{microtype}      
\usepackage{lipsum}
\usepackage{fancyhdr}       
\usepackage{graphicx}       
\usepackage{amsmath}
\usepackage{multirow}
\graphicspath{{media/}}     
\usepackage{algorithm}
\usepackage{algpseudocode}

\title{FedStein: Enhancing Multi-Domain Federated Learning Through James-Stein Estimator
}

\author{
  Sunny Gupta \\
  Koita Centre for Digital Health \\
  Indian Institute of Technology, Bombay \\
  Mumbai, Maharashtra 400076 \\
  \texttt{sunnygupta@iitb.ac.in} \\
  \And
  Nikita Jangid \\
  Department of Electrical Engineering \\
  Indian Institute of Technology, Bombay \\
  Mumbai, Maharashtra 400076 \\
  \texttt{21d070032@iitb.ac.in} \\
  \And
  Amit Sethi \\
  Department of Electrical Engineering \\
  Indian Institute of Technology, Bombay \\
  Mumbai, Maharashtra 400076 \\
  \texttt{asethi@iitb.ac.in} \\
}

\begin{document}
\maketitle

\begin{abstract}
Federated Learning (FL) facilitates data privacy by enabling collaborative in-situ training across decentralized clients. Despite its inherent advantages, FL faces significant challenges of performance and convergence when dealing with data that is not independently and identically distributed (non-i.i.d.). While previous research has primarily addressed the issue of skewed label distribution across clients, this study focuses on the less explored challenge of multi-domain FL, where client data originates from distinct domains with varying feature distributions. We introduce a novel method designed to address these challenges -- \textbf{FedStein}: Enhancing Multi-Domain \textbf{Fed}erated Learning Through the James-\textbf{Stein} Estimator. FedStein uniquely shares only the James-Stein (JS) estimates of batch normalization (BN) statistics across clients, while maintaining local BN parameters. The non-BN layer parameters are exchanged via standard FL techniques. Extensive experiments conducted across three datasets and multiple models demonstrate that FedStein surpasses existing methods such as FedAvg and FedBN, with accuracy improvements exceeding 14\% in certain domains leading to enhanced domain generalization. The code is available at \url{https://github.com/sunnyinAI/FedStein}
\end{abstract}


\section{Introduction}
Federated learning (FL) represents a transformative paradigm in machine learning, enabling collaborative modelling across decentralized devices while maintaining local data privacy. Unlike traditional centralized methods, FL conducts model training directly on individual devices, transmitting only model updates instead of raw data. This approach not only preserves data privacy, but also aligns with stringent data governance standards, making FL particularly appealing in a variety of domains, including healthcare \cite{li2019privacy, bernecker2022fednorm}, mobile devices \cite{hard2018federated, paulik2021federated}, and autonomous vehicles \cite{zhang2021end, posner2021federated, nguyen2022deep}.
However, FL faces significant challenges when applied to data that are not independently and identically distributed (non-i.i.d.) between different clients \cite{li2020federated}. These challenges manifest themselves as performance degradation \cite{tan2024heterogeneity, hsieh2020non,zhao2018federated}, instability during training \cite{wang2023batch,zhuang2023optimizing,zhuang2020performance}, and biases in the resulting models.
Most research addressing non-i.i.d. challenges in FL has focused on skewed label distributions, where each client has a different distribution of labels \cite{chen2022calfat, wang2020federated, hsieh2020non,li2020federated}. While this focus is crucial, it often overlooks a critical aspect of real-world FL applications: multi-domain federated learning. In multi-domain FL, client data come from diverse domains, each characterized by unique feature distributions rather than merely differing in label distributions. For example, autonomous vehicles may collect data under various weather conditions or at different times of the day, leading to domain gaps in the images captured by a single client \cite{yu2020bdd100k,cordts2016cityscapes}. Similarly, image data from different institutions in healthcare can show significant variation due to differences in equipment and protocols \cite{bernecker2022fednorm}.

\begin{figure*}[ht]
    \centering
\includegraphics[width=13cm,height=6.5cm]{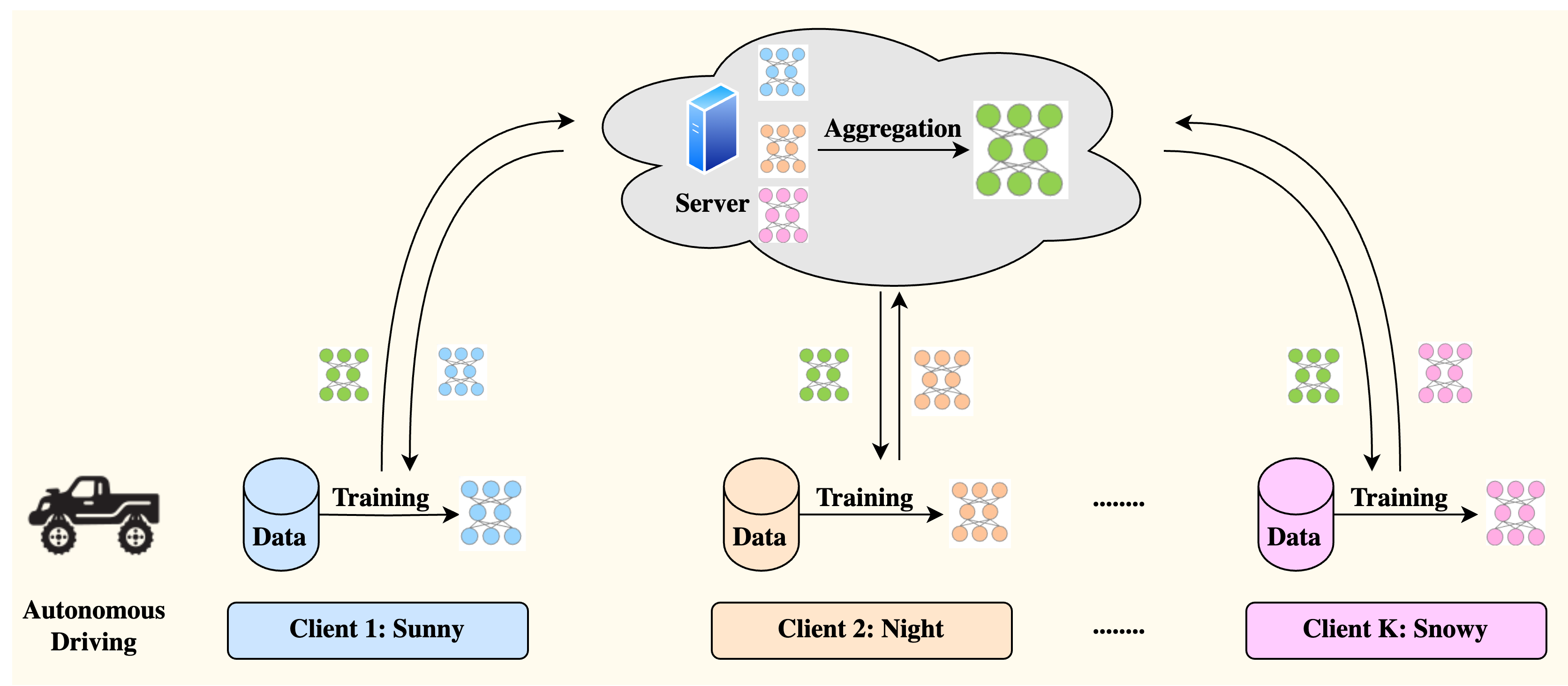} 
    \caption{We focus on multi-domain federated learning, where each client possesses data from a specific domain. This framework is highly relevant and useful in real-world applications. For example, autonomous vehicles in different regions gather images under various weather conditions.}
    \label{multidomainfl}
\end{figure*}
 
The existing solutions fail to effectively address the challenges posed by multi-domain Federated Learning (FL). One notable attempt, FedBN \cite{li2021fedbn}, seeks to mitigate this issue by retaining Batch Normalization (BN) parameters and statistics \cite{ioffe2015batch} locally on each client. However, FedBN is primarily suited for cross-silo FL scenarios, where clients are typically stable organizations, such as healthcare institutions, capable of maintaining state \cite{karimireddy2020scaffold} and consistently participating in every training round. This requirement for clients to be stateful means that they must preserve BN-related information across training rounds. By contrast, FedBN is less suitable for cross-device FL scenarios, where clients are stateless and only a subset participates in training. 

Moreover, BN assumes that the training data originates from a single distribution, ensuring that the mean and variance computed from each mini-batch are representative of the entire dataset \cite{ioffe2015batch}. 

Although alternative normalization techniques, such as Layer Norm \cite{ba2016layer} and Group Norm \cite{wu2018group}, have been proposed, their applicability to multi-domain FL has not been thoroughly explored, and they come with limitations such as increased computational overhead during inference.

This paper proposes an approach to address the challenges inherent in multi-domain Federated Learning (FL). Given the difficulties BN faces in handling multi-domain data and the limitations of alternative normalization techniques, we explore a critical question: Is normalization truly indispensable for learning a robust global model in multi-domain FL? Recent research on normalization techniques using the James-Stein Estimator \cite{khoshsirat2024improving} suggests that models can achieve better performance than standard BN by countering differences between the statistics between batches \cite{ioffe2015batch}. Building on this insight, we investigate the potential of this methodology within the context of multi-domain FL.

FedStein adheres to the FedAvg \cite{mcmahan2017communication} protocols for server aggregation and client training. However, unlike existing methods, FedStein selectively aggregates only the essential James-Stein (JS) estimates of Batch Normalization (BN) statistics -- namely the mean (\(\mu_{\text{JS}}\)) and variance (\(\sigma_{\text{JS}}^2\)) -- while eliminating the need to synchronize the BN parameters (\(\gamma\) and \(\beta\)). The parameters of non-BN layers are disseminated conventionally, thereby maintaining the integrity of the model's performance across diverse domains. 

We conducted extensive experiments in three datasets and confirmed that FedStein outperforms state-of-the-art methods in all cases. The global model trained using FedStein achieves an improvement of more than 14\% on certain domains compared to the personalized models generated by FedBN \cite{li2021fedbn}.

Our contributions to this study are threefold:
\begin{enumerate}
    \item We formulate FedStein, a new FL approach to address domain discrepancies among clients in multi-domain Federated Learning (FL).
    
    \item We demonstrate that, unlike traditional methods, FedStein effectively tackles the challenges posed by non-identically distributed data in FL, ensuring a more resilient and accurate model across varied datasets.
    
    \item We show that FedStein is versatile in offering robust support for cross-silo Federated Learning (FL) environments by introducing JS estimates of BN Statistics. This approach further enhances privacy by not communicating local activation statistics, thereby revealing less sensitive information.
\end{enumerate}
\section{Related Work}

\textbf{Batch Normalization:} Batch Normalization \cite{ioffe2015batch} process involves normalizing the inputs of each layer by computing the mean and variance from a mini-batch, followed by scaling and shifting through learnable parameters. Specifically, given a batch of inputs $\{x_1, x_2, \dots, x_m\}$, BN calculates the mean $\mu_B$ and variance $\sigma_B^2$ as follows:
\vspace{-0.3em} 
\begin{equation}
\mu_B = \frac{1}{m} \sum_{i=1}^{m} x_i
\end{equation}
\vspace{-0.3em} 
\begin{equation}
\sigma_B^2 = \frac{1}{m} \sum_{i=1}^{m} (x_i - \mu_B)^2
\end{equation}

The normalized output is then computed as:
\vspace{-0.3em} 
\begin{equation}
\hat{x}_i = \frac{x_i - \mu_B}{\sqrt{\sigma_B^2 + \epsilon}},
\end{equation}

To provide the network with the flexibility to adjust the normalization, BN incorporates learnable scaling and shifting parameters, $\gamma$ and $\beta$, applied as follows:
\vspace{-0.3em} 
\begin{equation}
y_i = \gamma \hat{x}_i + \beta.
\end{equation}
where $\mu$ and $\sigma^2$, hereafter denoted as BN statistics, are calculated as the running means and variances, respectively, of each channel computed across both spatial and batch dimensions, and $\gamma$ and $\beta$ are learned affine renormalization parameters, and where all computations are performed along the channel axis. The term $\epsilon$ is a small positive constant added for numerical stability.

\subsection{Federated Learning with Batch Normalization}
In centralized training paradigms, batch normalization (BN) has faced significant challenges when modelling statistics across multiple domains. This limitation has spurred the development of domain-specific BN techniques designed to better accommodate variability in data distributions \cite{li2016revisiting, chang2019domain}. These challenges are further exacerbated in the context of multi-domain Federated Learning (FL), where deep neural networks (DNNs) that rely on BN may struggle to accurately capture the statistical characteristics of diverse domains while attempting to train a unified global model.

A prominent issue addressed for non-i.i.d. Federated learning (FL) is the skewed label distribution, where label distributions vary significantly across clients. This disparity can lead to biased model performance and hinder generalization. To address this challenge, several strategies have been proposed, including specialized operations in BN to tailor the models to the unique data distributions of each client \cite{li2020federated, zhang2023client, chen2022fedcor}. For example, SiloBN \cite{andreux2020siloed} retains BN statistics locally on each client, ensuring that the normalization process is customized to the specific data distribution of the client. Similarly, FixBN \cite{zhong2023fixbn} mitigates the problem by training BN statistics during the early stages and subsequently freezing them to maintain consistency. 

In contrast, multi-domain FL has received comparatively less attention \cite{chen2018multi, shen2022multi}. To address the unique challenges posed by multi-domain FL approaches such as FedBN \cite{li2021fedbn} and FedNorm \cite{bernecker2022fednorm} have been developed. These methods retain BN layers locally on clients while aggregating only the remaining model parameters. Similarly, PartialFed \cite{sun2021partialfed} preserves model initialization strategies on clients, leveraging these strategies to load models in subsequent training rounds.


\subsection{Alternative Normalization Methods}
Despite its widespread effectiveness, BN encounters several limitations in certain scenarios. For example, BN may struggle to accurately capture the statistical properties of training data originating from multiple domains \cite{li2016revisiting, chang2019domain}. To address these limitations, researchers have proposed alternative normalization techniques, such as Group Normalization (GN) \cite{wu2018group} and Layer Normalization (LN) \cite{ba2016layer}. Although these methods alleviate some of the constraints associated with BN, such as its dependence on batch size, they introduce their own set of challenges. For example, both GN and LN require additional computational overhead during inference, which can limit their practicality, particularly in edge-to-edge deployment scenarios where computational resources are constrained.
Recent studies have further highlighted that Batch Normalization (BN) may not perform optimally in Federated Learning (FL) environments, especially under non-i.i.d. data conditions \cite{hsieh2020non}. 
This suboptimal performance is primarily because of external covariate shifts \cite{du2022does} and the mismatch between local and global statistics \cite{wang2023fedtan}. In response to these challenges, alternative normalization techniques such as Group Normalization (GN) \cite{hsieh2020non, casella2023gn} and Layer Normalization (LN) \cite{du2022does, casella2023gn} have been explored. However, these alternatives are not without their drawbacks. GN and LN inherit some of the limitations observed in centralized training, including increased computational complexity and sensitivity to specific hyperparameters. 

\subsection{Normalization with James-Stein Estimator}
One notable concern about Equation 1 lies in the estimation of the mean and variance. The conventional approach suggests independently calculating the mean and variance using “usual estimators”. For batch normalization, the estimators are given as follows:
\vspace{-0.3em}
\begin{equation}
\mathbb{E}[x_i] = \frac{1}{n} \sum_{j=1}^{n} x_{i,j},
\end{equation}
\vspace{-0.3em}
\begin{equation}
\text{Var}[x_i] = \frac{1}{n} \sum_{j=1}^{n} (x_{i,j} - \mathbb{E}[x_i])^2.
\end{equation}
Given that all the features contribute to a shared loss function, according to Stein’s paradox \cite{wikipedia2023jamesstein}, these estimators are inadmissible when $c \geq 3$. Notably, in computer vision networks, it is consistently observed that $c \geq 3$. To address this, a novel method was adopted by \cite{khoshsirat2024improving} to adopt admissible shrinkage estimators, which effectively enhance the estimation of the mean and variance in normalization layers.

Let $X = \{x_1, x_2, \dots, x_c\}$ with unknown means $\theta = \{\theta_1, \theta_2, \dots, \theta_c\}$ and estimates $\hat{\theta} = \{\hat{\theta}_1, \hat{\theta}_2, \dots, \hat{\theta}_c\}$. The basic formula for the James-Stein estimator is:
\vspace{-0.3em}
\begin{equation}
\hat{\theta}_{JS} = \hat{\theta} + s(\mu_{\hat{\theta}} - \hat{\theta}),
\end{equation}
where $\mu_{\hat{\theta}} - \hat{\theta}$ is the difference between the total mean (average of averages) and each estimated mean, and $s$ is a shrinking factor. Among the numerous perspectives that motivate the James-Stein estimator, the empirical Bayes perspective \cite{efron1975data} is the most insightful. Taking a Gaussian prior on the unknown means leads us to the following formula \cite{james1992estimation}:
\vspace{-0.3em}
\begin{equation}
\hat{\theta}_{JS} = \left(1 - \frac{(c - 2)\sigma^2}{\|\hat{\theta} - v\|^2_2}\right) (\hat{\theta} - v) + v,
\end{equation}
where $\|\cdot\|_2$ denotes the $L_2$ norm of the argument, $\sigma^2$ is the variance, $v$ is an arbitrarily fixed vector that shows the shrinkage direction, and $c \geq 3$. Setting $v = 0$ results in the following:
\vspace{-0.3em}
\begin{equation}
\hat{\theta}_{JS} = \left(1 - \frac{(c - 2)\sigma^2}{\|\hat{\theta}\|^2_2}\right)\hat{\theta}.
\end{equation}
The above estimator shrinks the estimates towards the origin $0$. 
Using Equation 6 helps to mitigate the 'mean shift' problem \cite{brock2021high, brock2021characterizing}. To incorporate this equation into normalization layers, we replace the $\hat{\theta}$ in Equation 6 with the estimated mean and variance derived from the original method. By applying the James-Stein estimator to these estimated statistics, the additional computational overhead is minimal and can be considered negligible \cite{khoshsirat2024improving}. Thus, the James-Stein estimator is utilized for both the mean and variance within the normalization layers. In the context of batch normalization, $\mathbb{E}[x]$ and $\text{Var}[x]$ are vectors of length $c$ (where $c$ represents the number of channels in the batch). These vectors can be directly substituted in place of $\hat{\theta}$.

More recently, \cite{khoshsirat2024improving} has shown that the mean and variance estimators commonly used in normalization layers are inadmissible. They introduced a novel approach that employs the James-Stein estimator \cite{stein1956inadmissibility}to enhance the estimation of mean and variance within these layers. This improved normalization technique consistently yields superior accuracy across a variety of tasks without imposing additional computational burdens. Their method achieves competitive results compared to Batch Normalization (BN) on prominent architectures such as ResNet \cite{he2016deep}, EfficientNet \cite{tan2019efficientnet}, and Swin Transformer v2 \cite{liu2022swin}.

\section{Methodology}
Batch Normalization offers several key advantages, including the mitigation of internal covariate shift, stabilization of the training process, and acceleration of convergence \cite{santurkar2018does}. BN also reduces the number of iterations required to reach convergence, thereby improving overall performance \cite{khoshsirat2023empowering}. Moreover, it enhances robustness to variations in hyperparameters \cite{bjorck2018understanding} and contributes to a smoother optimization landscape \cite{santurkar2018does}. However, the effectiveness of BN is predicated on the assumption that the training data is homogeneous across the dataset, ensuring that the mean ($\mu$) and variance ($\sigma$) computed from each mini-batch are representative of the overall data distribution \cite{ioffe2015batch}.
\vspace{-0.3em}

\begin{figure*}[htbp]
    \centering
    \includegraphics[width = 12cm,height=4 cm]{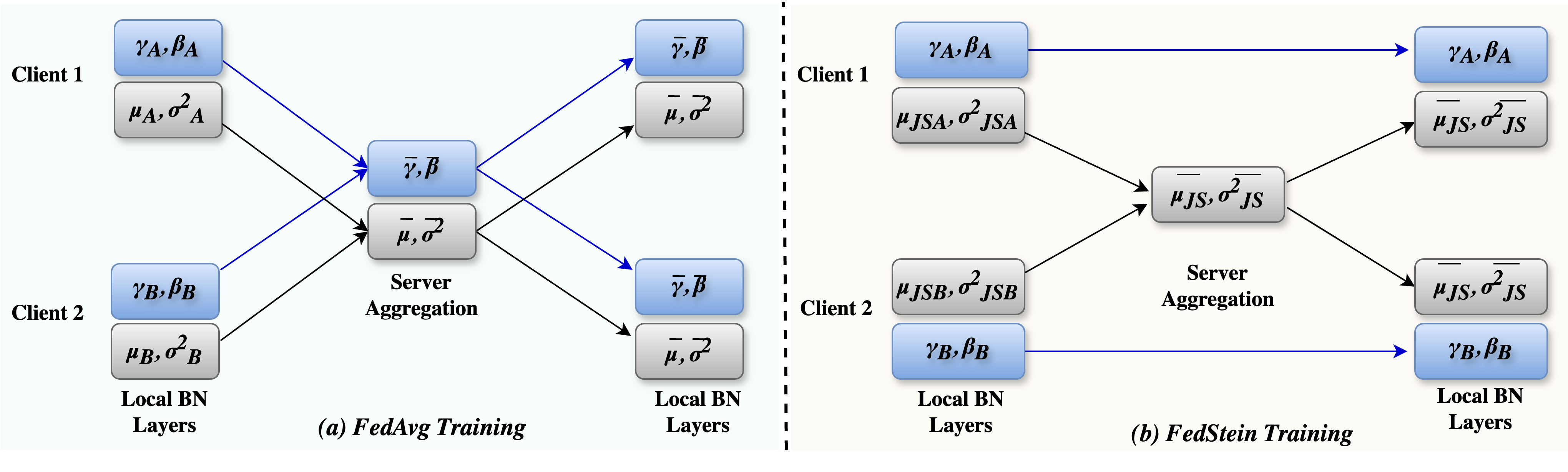}
    \caption{An illustration of two different approaches to multi-centre training of BN layers. This description follows the definitions provided in Eqn 3. Computation flows from left to right. (a) In \textbf{FedAvg}, both BN parameters and BN statistics are aggregated into one server, and (b) in \textbf{FedStein}, BN parameters are removed, and the JS norms of the BN statistics are aggregated into the server. Non-BN layers are shared in both methods.}
    \label{fedstein}
\end{figure*}

\begin{algorithm}[htbp]
\caption{Aggregated Batch Normalization with James-Stein Estimator}
\label{alg:AggregatedJSNorm}
\begin{algorithmic}[1]
\small 

\State \textbf{Notations:} 
\State $k$: Index for user;
\State $l$: Index for neural network layer;
\State $w^{(l)}_{0,k}$: Initialized model parameters;
\State $E$: Local update frequency;
\State $T$: Total rounds;
\State $K$: Number of users;
\Require Batch of inputs $\{x_1, x_2, \dots, x_m\}$, small constant $\epsilon$, learnable parameters $\gamma$, $\beta$, dimensionality $c$
\Ensure Normalized outputs $\{y_1, y_2, \dots, y_m\}$

\For{each round $t = 1$ to $T$}
    \For{each user $k$ and each layer $l$}
        \State Update model parameters
        \[
        w^{(l)}_{t+1,k} \leftarrow \text{SGD}(w^{(l)}_{t,k})
        \]
    \EndFor

    \If{mod($t$, $E$) = 0}
        \For{each user $k$ and each layer $l$}
            \If{layer $l$ is not BatchNorm}
                \State Aggregate the model parameters:
                \[
                w^{(l)}_{t+1,k} \leftarrow \frac{1}{K} \sum_{k=1}^{K} w^{(l)}_{t+1,k}
                \]
            \Else
                \State \textbf{(Do not aggregate BN parameters $\gamma$ and $\beta$)}
                \State Aggregate the BatchNorm statistics across users:
                
                \State Aggregate the batch mean:
                \[
                \mu_{B, \text{agg}} = \frac{1}{K} \sum_{k=1}^{K} \mu_B^{(k)}
                \]
                \State Aggregate the batch variance:
                \[
                \sigma_{B, \text{agg}}^2 = \frac{1}{K} \sum_{k=1}^{K} \sigma_B^{2(k)}
                \]
                
                \State Adjust the aggregated mean using the James-Stein Estimator:
                \[
                \mu_{JS, \text{agg}} = \left(1 - \frac{(c-2)\sigma_{\mu_B, \text{agg}}^2}{\|\mu_{B, \text{agg}}\|^2}\right) \mu_{B, \text{agg}}
                \]
                \State Adjust the aggregated variance using the James-Stein Estimator:
                \[
                \sigma_{JS, \text{agg}}^2 = \left(1 - \frac{(c-2)\sigma_{\sigma_B, \text{agg}}^2}{\|\sigma_{B, \text{agg}}^2\|^2}\right) \sigma_{B, \text{agg}}^2
                \]
                
                \For{each input $x_i$ in the batch}
                    \State Normalize the input using aggregated statistics:
                    \[
                    \hat{x}_i = \frac{x_i - \mu_{JS, \text{agg}}}{\sqrt{\sigma_{JS, \text{agg}}^2 + \epsilon}}
                    \]
                    \State Apply scaling and shifting:
                    \[
                    y_i = \gamma \hat{x}_i + \beta
                    \]
                \EndFor
            \EndIf
        \EndFor
    \EndIf
\EndFor

\Return Normalized outputs $\{y_1, y_2, \dots, y_m\}$

\end{algorithmic}
\end{algorithm}

\subsection{James-Stein Estimator}
Estimating the mean of a multivariate normal distribution is a fundamental problem in statistics. Typically, the sample mean is employed, which also serves as the maximum-likelihood estimator. However, the James-Stein (JS) estimator, despite being biased, is utilized for estimating the mean of \(c\) correlated Gaussian-distributed random vectors with unknown means. The development of the JS estimator is rooted in two pivotal papers, with the initial version introduced by Charles Stein in 1956 \cite{stein1956inadmissibility}. Stein's work led to the surprising revelation that the standard mean estimate is admissible when \(c \leq 2\) but becomes inadmissible when \(c \geq 3\). This breakthrough suggested an improvement by shrinking the sample means towards a central vector of means, a concept commonly referred to as Stein's paradox or Stein's example \cite{efron1977stein}.

\subsection{Applying James-Stein Estimation to Batch Normalization} 
Implementing James-Stein Normalization (JSNorm) by incorporating the James-Stein estimator into the standard Batch Normalization (BN) framework to enhance the robustness of BN in federated learning scenarios, where data distributions can be heterogeneous and high-dimensional, the batch mean $\mu_B$ and variance $\sigma_B^2$ are adjusted using the James-Stein estimator, resulting in the modified statistics as per Algorithm~\ref{alg:AggregatedJSNorm}:






\subsection{JSNorm in Federated Learning} 
While Batch Normalization (BN) layers are integral components of many modern neural network architectures \cite{he2016deep, hu2018squeeze, tan2019efficientnet}, their application in federated learning settings has not been thoroughly investigated and is often overlooked or omitted entirely \cite{mcmahan2017communication}. The naïve implementation of FedAvg, for instance, does not differentiate between the local activation statistics ($\mu$, $\sigma^2$) and the trained renormalization parameters ($\gamma$, $\beta$), leading to a straightforward aggregation of both at each federated round, as illustrated in Figure 2 (left). This simplistic approach serves as a baseline for using FedAvg in networks that include BN layers.
Moreover, BN layers can serve a dual purpose by distinguishing between local and domain-invariant information. Specifically, the BN statistics and the learned BN parameters fulfil different roles \cite{li2016revisiting}: the former encapsulates local domain-specific information, while the latter can be transferred across different domains. 
In traditional federated learning approaches, all Batch Normalization (BN) parameters are typically treated equally during aggregation. However, this overlooks the distinct roles played by BN statistics (\(\mu\), \(\sigma^2\)) and learned parameters (\(\gamma\), \(\beta\)). 
We propose a new method called FedStein, which differentiates between these roles by sharing only the James-Stein (JS) estimates of BN statistics across different federated centres while keeping the learned BN parameters local to each centre. The parameters of non-BN layers are shared using the standard federated learning approach. This method is illustrated in Figure 2 (right). 
By synchronizing the JS estimates of BN statistics, FedStein enables the federated training of a model that is more robust to the heterogeneity present across different centres, thereby improving the overall model performance in multi-domain federated learning scenarios. By incorporating the James-Stein adjustment, JSNorm provides a more robust estimation of these statistics by guiding the sample means toward a more centralized mean vector, resulting in improved model performance across clients. This adjustment is particularly advantageous in scenarios involving high-dimensional data or small batch sizes, where standard Batch Normalization (BN) might otherwise face significant challenges. The results show that our improved normalization layers consistently deliver superior accuracy, without incurring any additional computational overhead.

\section{Experimental Evaluation}
\subsection{Results on Cross-silo Federated Learning}
\begin{table*}[ht!]
\centering
\scriptsize 
\setlength{\tabcolsep}{3pt} 
\begin{tabular}{llccccccccc}
\toprule
\textbf{Dataset} & \textbf{Domains} & \textbf{SingleSet} & \textbf{FedAvg} & \textbf{FedProx} & \textbf{+GN$^a$} & \textbf{+LN$^b$} & \textbf{SiloBN} & \textbf{FixBN} & \textbf{FedBN} & \textbf{FedStein (Ours)}\\
\midrule
\multirow{6}{*}{\textbf{Digit-Five}} 
& MNIST       & 94.4 & 96.2 & 96.4 & 96.4 & 96.4 & 96.2 & 96.3 & 96.3 & \textbf{96.8} \\
& SVHN        & 67.1 & 71.6 & 71.0 & \textbf{76.9} & 75.2 & 71.3 & 71.3 & 71.1 & 75.4 \\
& USPS        & 95.4 & 96.3 & 96.1 & 96.6 & 96.4 & 96.0 & 96.1 & 96.6 & \textbf{97.1} \\
& SynthDigits & 80.3 & 86.0 & 85.9 & 85.6 & 85.6 & 86.0 & 84.8 & \textbf{86.8} & 86.6 \\
& MNIST-M     & 77.0 & 82.5 & 83.1 & 83.7 & 82.2 & 83.1 & 83.0 & 78.6 & \textbf{84.1} \\
& \textbf{Average} & 83.1 & 86.5 & 86.5 & 87.8 & 87.1 & 86.5 & 86.5 & 87.2 & \textbf{88.1} \\
\midrule
\multirow{5}{*}{\textbf{Caltech-10}} 
& Amazon      & 54.5 & 61.8 & 59.9 & 60.8 & 55.0 & 60.8 & 59.2 & 63.2 & \textbf{64.1} \\
& Caltech     & 40.2 & 44.9 & 44.0 & 50.8 & 41.3 & 44.4 & 44.0 & 45.3 & \textbf{47.6} \\
& DSLR        & 81.3 & 77.1 & 76.0 & 88.5 & 79.2 & 76.0 & 79.2 & 83.1 & \textbf{83.4} \\
& Webcam      & 89.3 & 81.4 & 80.8 & 83.6 & 71.8 & 81.9 & 79.6 & 85.6 & \textbf{93.2} \\
& \textbf{Average} & 66.3 & 66.3 & 65.2 & 70.9 & 61.8 & 65.8 & 65.5 & 69.2 & \textbf{72.1} \\
\midrule
\multirow{7}{*}{\textbf{DomainNet}} 
& Clipart     & 42.7 & 48.9 & 51.1 & 45.4 & 42.7 & 51.8 & 49.2 & 51.2 & \textbf{58.4} \\
& Infograph   & 24.0 & 26.5 & 24.1 & 21.1 & 23.6 & 25.0 & 24.5 & 26.8 & \textbf{27.6} \\
& Painting    & 34.2 & 37.7 & 37.3 & 35.4 & 35.3 & 36.4 & 38.2 & 41.5 & \textbf{51.5} \\
& Quickdraw   & \textbf{70.9} & 44.5 & 46.1 & 57.2 & 46.0 & 45.9 & 46.3 & 64.8 & 55.4 \\
& Real        & 51.2 & 46.8 & 45.5 & 50.7 & 43.9 & 47.7 & 46.2 & 55.2 & \textbf{61.5} \\
& Sketch      & 33.5 & 35.7 & 37.5 & 36.5 & 28.9 & 38.0 & 37.4 & 39.6 & \textbf{40.3} \\
& \textbf{Average} & 42.9 & 40.0 & 40.2 & 41.1 & 36.7 & 40.8 & 40.3 & 45.8 & \textbf{49.1} \\
\bottomrule
\end{tabular}
\caption{Comparison of testing accuracy (\%) across different methods on three datasets. The proposed FedStein method consistently outperforms existing approaches in the majority of domains, achieving the highest average testing accuracy across all datasets.}
\label{tab:crosssilo}
\footnotesize{$^a$+GN means FedAvg+GN, $^b$+LN means FedAvg+LN}
\end{table*}

Table \ref{tab:crosssilo} presents a comprehensive comparison of various methods under cross-silo federated learning (FL) on the Digits-Five, Office-Caltech-10, and DomainNet datasets. The proposed FedStein method consistently outperformed state-of-the-art approaches across most domains and datasets. Notably, FedProx, which incorporates a proximal term into FedAvg, exhibits performance similar to FedAvg. These two methods surpass SingleSet on the Digits-Five dataset but may demonstrate inferior performance compared to SingleSet in certain domains on the more challenging Office-Caltech-10 and DomainNet datasets. \par
\textbf{Datasets}: We conducted a series of experiments on multi-domain federated learning (FL) using three datasets: Digits-Five \cite{li2021fedbn}, Office-Caltech-10 \cite{gong2012geodesic}, and DomainNet \cite{peng2019moment}. The Digits-Five dataset encompasses five distinct sets of 28x28 pixel digit images: MNIST \cite{lecun1998gradient}, SVHN \cite{netzer2011reading}, USPS \cite{hull1994database}, SynthDigits \cite{ganin2015unsupervised}, and MNIST-M \cite{ganin2015unsupervised}, each representing a unique domain. Office-Caltech-10 comprises real-world object images from four domains: three sourced from the Office-31 dataset (WebCam, DSLR, and Amazon) \cite{saenko2010adapting}, and one from the Caltech-256 dataset (Caltech) \cite{griffin2007caltech}. DomainNet \cite{peng2019moment}, known for its complexity, includes large 244x244 pixel object images spanning six domains: Clipart, Infograph, Painting, Quickdraw, Real, and Sketch.

To simulate realistic scenarios with limited client data, we used a reduced version of the standard digit datasets from \cite{li2021fedbn}, limiting them to 7,438 training samples, divided uniformly among 20 clients to mimic a cross-device FL setup, with a total of 100 clients. For the DomainNet dataset, we focused on 10 classes, each containing 2,000 to 5,000 images, assigning each client images from a single domain to simulate multi-domain FL. \par
\textbf{Implementation Details}
We implemented FedStein using PyTorch \cite{paszke2017automatic} and ran experiments on a high-performance cluster with four NVIDIA A6000 GPUs. The models tested included a 6-layer CNN \cite{li2021fedbn} for Digits-Five, AlexNet \cite{krizhevsky2017imagenet} and ResNet-18 \cite{he2016deep} for Office-Caltech-10, and AlexNet for DomainNet. We used cross-entropy loss for classification error and optimized with SGD. Learning rates were fine-tuned between [0.001, and 0.1] for optimal performance.
 \par

\textbf{Performance Evaluation}
We evaluated the performance of our proposed FedStein method by benchmarking it against three distinct categories of approaches. The first category includes state-of-the-art methods, which feature advanced techniques for Batch Normalization (BN), such as SiloBN \cite{andreux2020siloed}, FedBN \cite{li2021fedbn}, and FixBN \cite{zhong2023fixbn}. The second category consists of baseline algorithms, encompassing widely recognized federated learning (FL) algorithms like FedProx \cite{li2020federated}, FedAvg \cite{mcmahan2017communication}, and SingleSet, where a model is trained independently on each client using only local data. Lastly, we examined alternative normalization methods, specifically exploring the performance of FedAvg in combination with Group Normalization (FedAvg+GN) and Layer Normalization (FedAvg+LN), where the BN layers were replaced with GN and LN layers, respectively.

SiloBN and FixBN show performance comparable to FedAvg in terms of average accuracy; however, they are not specifically designed for multi-domain FL, and consequently, they achieve only baseline results. In contrast, FedBN, explicitly designed for multi-domain FL, surpasses these methods in overall performance.

Furthermore, our findings indicate that simply replacing BN with GN (FedAvg+GN) can enhance the performance of FedAvg, as GN does not rely on batch-specific domain statistics. FedAvg+GN achieves results comparable to FedBN on the Digits-Five and Office-Caltech-10 datasets. Remarkably, our proposed FedStein method surpasses both FedAvg+GN and FedBN in terms of average accuracy across all datasets. While FedStein trails FedBN by less than 1\% in two domains, it outperforms FedBN by more than 10\% in specific domains. These results underscore the effectiveness of FedStein in cross-silo FL scenarios.

\subsection{Results on medical images.}
\begin{table}[ht]
\centering
\setlength{\abovecaptionskip}{6pt} 
\setlength{\belowcaptionskip}{10pt} 
\begin{tabular}{lcccccc}
\hline
\textbf{Methods} & \textbf{Center 1} & \textbf{Center 2} & \textbf{Center 3} & \textbf{Center 4} & \textbf{Center 5} & \textbf{Center 6} \\
\hline
FedAvg          & 0.40          & 0.21          & 0.37          & 0.42          & 0.39          & 0.43          \\
FedBN           & 0.31          & 0.38          & 0.43          & 0.39          & 0.30          & 0.36          \\
\textbf{FedStein (Ours)} & \textbf{0.52} & \textbf{0.57} & \textbf{0.65} & \textbf{0.56} & \textbf{0.46} & \textbf{0.62} \\
\hline
\end{tabular}
\caption{Evaluation on Fed-ISIC2019 dataset with medical images from six different centres. FedStein outperforms FedAvg and FedBN by a significant margin in all domains.}
\label{tab:isic}
\end{table}
As a further experiment to demonstrate the advantages of our proposed FedStein approach in Federated Learning (FL) across multi-domains, we extend our analysis to the diagnosis of skin lesions utilizing datasets from the ISIC2019 Challenge \cite{codella2018skin, combalia2019bcn20000} and the HAM10000 dataset \cite{tschandl2018ham10000}. There are four hospitals represented in these datasets, including one hospital employing three different imaging technologies. Like that described by Flamby \cite{ogier2022flamby}, we have organized the data into six centres: BCN, Vidir-molemax, Vidir-modern, Rosendahl, MSK, and Vienna-dias. Each centre's data is associated with its domain, ensuring that the images reflect the differences found between many healthcare institutions and imaging methods. \par
\textbf{Datasets}: The dataset comprises a total of 23,247 images of skin lesions, distributed as follows: BCN contains 9930 training samples and 2483 testing samples; Vidir-molemax includes 3163 training samples and 791 testing samples; Vidir-modern has 2691 training samples and 672 testing samples; Rosendahl provides 1807 training samples and 452 testing samples; MSK comprises 655 training samples and 164 testing samples; and Vienna-dias includes 351 training samples and 88 testing samples.


\par
\textbf{Implementation Details}
We conducted experiments using ResNet-18 \cite{he2016deep} without pre-training, with local epochs \(E = 1\), batch size \(B = 64\), and 50 rounds. We used the SGD optimizer with learning rates \(\eta = 0.005\) for FedAvg and FedStein, and \(\eta = 0.001\) for FedBN, tuning \(\eta\) from \{0.001, 0.005, 0.01, 0.05\}. Following the implementation in Flamby, we employed weighted focal loss \cite{lin2017focal} and data augmentations. Table \ref{tab:isic} shows that FedStein consistently outperforms FedAvg and FedBN across six healthcare domains, demonstrating its potential for multi-domain healthcare scenarios where data is scarce and fragmented.

\subsection{Results on Domain Adaptation and Generalization}
\begin{table*}[ht]
\centering
\setlength{\abovecaptionskip}{6pt} 
\setlength{\belowcaptionskip}{10pt} 
\resizebox{\textwidth}{!}{ 
\begin{tabular}{lcccc}
\toprule
\textbf{Methods} & \textbf{Amazon (Seen)} & \textbf{Caltech (Seen)} & \textbf{DSLR (Seen)} & \textbf{WebCam (Unseen)} \\ 
\midrule
FedAvg & 59.9 & 43.2 & 60.1 & 53.2 \\ 
FedBN & 64.8 & 49.0 & 77.9 & 60.9 \\ 
\textbf{FedStein (Ours)} & \textbf{66.9} & \textbf{53.7} & \textbf{92.3} & \textbf{69.8} \\ 
\bottomrule
\end{tabular}
}
\caption{Comparison of methods on domain generalization using the Office-Caltech-10 dataset.}
\label{tab:domaingeneralization}
\end{table*}

The experiments for these results were performed on \textit{Office-Caltech10} dataset, with \textit{Amazon (A)}, \textit{Caltech (C)}, and \textit{DSLR (D)} designated as the ``seen'' domains during the training phase. The \textit{WebCam (W)} domain, conversely, was exclusively reserved for the evaluation phase, functioning as the ``unseen'' domain in a zero-shot evaluation context. To rigorously assess the model's performance, we utilized client local models to evaluate the seen domains, while the server global model was applied to test the unseen domain. 
The results, as documented in Table \ref{tab:domaingeneralization}, compellingly illustrate that FedStein not only surpasses other models in performance across the seen domains but also demonstrates superior generalization capabilities on the unseen domain. These findings underscore the significant advantage of FedStein, particularly in the context of domain generalization.

\section{Conclusion}
In this paper, we introduced FedStein, a novel approach for enhancing multi-domain Federated Learning (FL) that relies on the James-Stein (JS) estimation of BN statistics across clients. This method effectively addresses feature shifts in non-i.i.d. data, particularly in multi-domain scenarios where data characteristics can vary significantly across clients. Through extensive experimentation on diverse federated datasets, we demonstrate that FedStein significantly enhances both convergence behaviour and model performance in non-i.i.d. settings. Privacy remains a critical concern in federated learning, and retaining the BN parameters locally at each client in FedStein adds a layer of security, making it more challenging to compromise local data. Our extensive experiments across three multi-domain datasets and models demonstrate that FedStein consistently surpasses state-of-the-art techniques, proving its versatility in both cross-silo and domain generalization scenarios. Future work could explore this method further by evaluating it across a broader range of datasets and model architectures, especially in the context of medical imaging.







\bibliographystyle{unsrt}  
\bibliography{references}
\end{document}